\documentclass[sigconf]{acmart}

\AtBeginDocument{%
  \providecommand\BibTeX{{%
    \normalfont B\kern-0.5em{\scshape i\kern-0.25em b}\kern-0.8em\TeX}}}

\copyrightyear{2019}
\acmYear{2019}
\setcopyright{acmcopyright}
\acmPrice{15.00}
\acmDOI{10.1145/3338533.3366556}
\acmISBN{978-1-4503-6841-4/19/12}

\usepackage{bm}

\begin{document}

\title{Robust Visual Tracking via Statistical Positive Sample Generation and Gradient Aware Learning}

 \author{Lijian Lin, Haosheng Chen, Yanjie Liang, Yan Yan, Hanzi Wang}
 \authornote{Corresponding author.}
  \affiliation{%
   \institution{\small{Fujian Key Laboratory of Sensing and Computing for Smart City, School of Informatics, Xiamen University, Xiamen, China}}
 }
 \email{ljlin@stu.xmu.edu.cn, haoshengchen@stu.xmu.edu.cn, yanjieliang@yeah.net, yanyan@xmu.edu.cn, hanzi.wang@xmu.edu.cn}

\begin{abstract}
In recent years, Convolutional Neural Network (CNN) based trackers have achieved state-of-the-art performance on multiple benchmark datasets. Most of these trackers train a binary classifier to distinguish the target from its background. However, they suffer from two limitations. Firstly, these trackers cannot effectively handle significant appearance variations due to the limited number of positive samples. Secondly, there exists a significant imbalance of gradient contributions between easy and hard samples, where the easy samples usually dominate the computation of gradient. In this paper, we propose a robust tracking method via Statistical Positive sample generation and Gradient Aware learning (SPGA) to address the above two limitations. To enrich the diversity of positive samples, we present an effective and efficient statistical positive sample generation algorithm to generate positive samples in the feature space. Furthermore, to handle the issue of imbalance between easy and hard samples, we propose a gradient sensitive loss to harmonize the gradient contributions between easy and hard samples. Extensive experiments on three challenging benchmark datasets including OTB50, OTB100 and VOT2016 demonstrate that the proposed SPGA performs favorably against several state-of-the-art trackers.
\end{abstract}

\ccsdesc[500]{Computing methodologies~Computer vision}
\ccsdesc[500]{Computing methodologies~Tracking}

\keywords{Visual Tracking, Sample Imbalance, Positive Sample Generation, Gradient Sensitive Loss}


\maketitle

\section{Introduction}

Visual tracking is one of the fundamental problems in multimedia and computer vision with a variety of applications such as video surveillance and autonomous driving \cite{yilmaz2006object, smeulders2013visual}. It aims to estimate the trajectory of a target in a video sequence, given only the bounding box of the target at the first frame. Despite the great progress during the past few decades, visual tracking remains a challenging task due to various factors such as deformation, occlusion, fast motion.

\begin{figure}
\setlength{\abovecaptionskip}{4pt}
\centering
  \includegraphics[width=0.322\linewidth, height=0.14\linewidth]{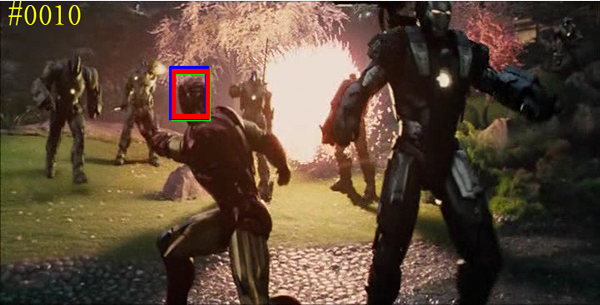}
  \includegraphics[width=0.322\linewidth, height=0.14\linewidth]{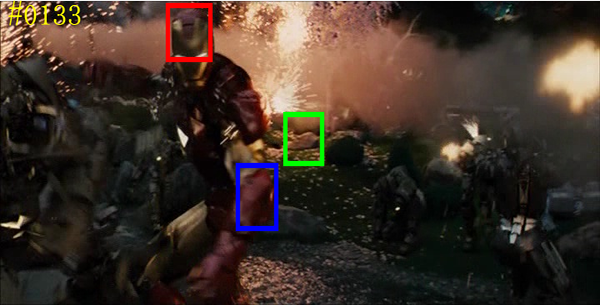}
  \includegraphics[width=0.322\linewidth, height=0.14\linewidth]{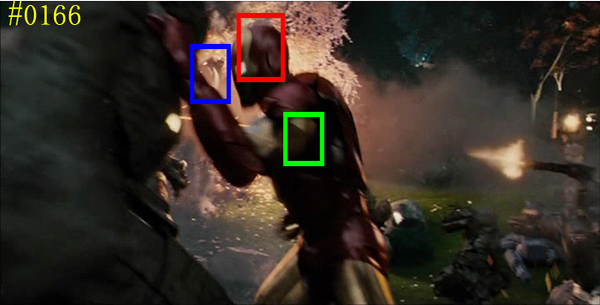}
  \includegraphics[width=0.322\linewidth, height=0.14\linewidth]{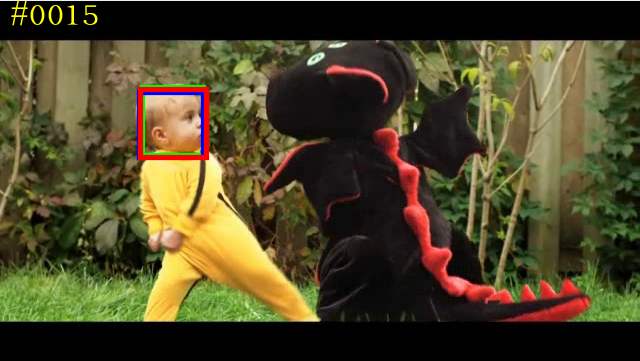}
  \includegraphics[width=0.322\linewidth, height=0.14\linewidth]{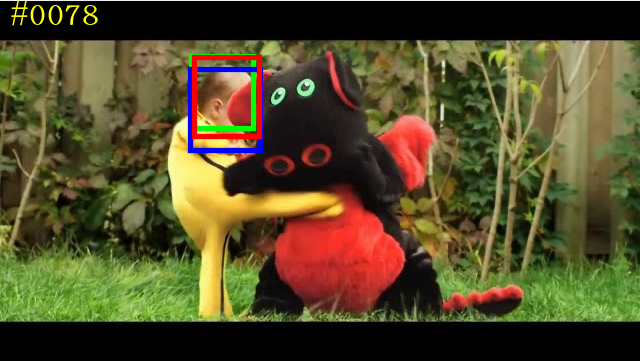}
  \includegraphics[width=0.322\linewidth, height=0.14\linewidth]{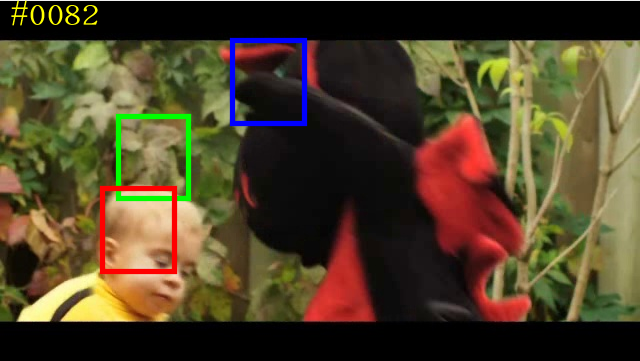}
  \includegraphics[width=0.322\linewidth, height=0.14\linewidth]{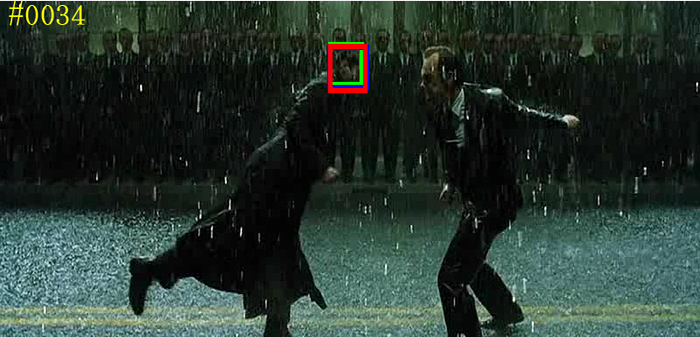}
  \includegraphics[width=0.322\linewidth, height=0.14\linewidth]{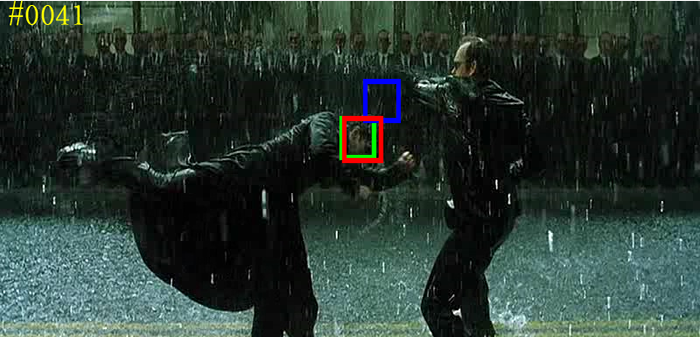}
  \includegraphics[width=0.322\linewidth, height=0.14\linewidth]{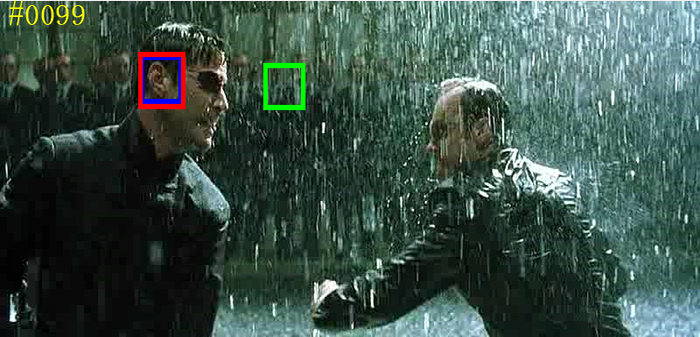}
  \includegraphics[width = \linewidth]{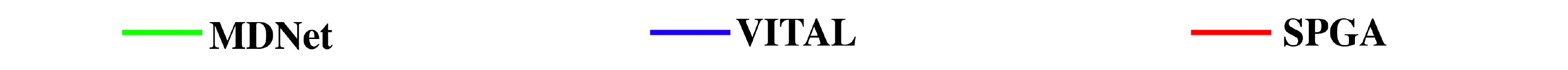} 
  \caption{Comparisons between our SPGA and two state-of-the-art trackers including MDNet \cite{MDNet} and VITAL \cite{vital}. Our SPGA performs favorably against these two trackers.}
  \label{fig1}

\end{figure}

Recently, Convolutional Neural Networks (CNNs) have gained great success in visual tracking. The state-of-the-art CNN based trackers usually pre-train their networks on large-scale datasets and fine-tune their networks using the samples collected during the tracking process. 
Despite the favorable performance of these trackers, they suffer from two limitations. On one hand, the appearance of the target varies frame-by-frame (caused by occlusion, deformation, motion blur, etc.) in the whole video sequence. The limited positive samples collected from previous frames make the trackers difficult to handle time-varying appearance changes in the subsequent frames, leading to the inferior tracking performance. On the other hand,
in visual tracking, positive samples are highly overlapped with the target, and a large portion of negative samples belong to the less informative background.
Thus most of the training samples are easy samples. As a consequence, the overall gradient in training a CNN based tracker is dominated by these easy samples, leading to severe imbalance of gradient contributions between easy and hard samples. Therefore, the limited positive sample problem and the imbalance problem need to be tackled.

To handle the problem of limited positive samples, recent trackers generally employ data augmentation strategies (e.g., data augmented by geometric or appearance transformations \cite{UPDT}, the Generative Adversarial Networks (GANs) \cite{vital}). 
Despite the promising performance of the GAN based data augmentation strategies, they suffer from high computational complexity and instability in training. Meanwhile, for the imbalance of gradient contributions between easy and hard samples, recent trackers usually resort to designing specific loss functions \cite{vital} or relying on hard negative mining mechanisms \cite{MDNet}. The designed loss functions down-weight the contributions of easy samples and emphasize the hard samples. These loss functions, however, usually ignore the distribution of the training samples in a mini-batch, thus leading to the inferior tracking performance.

In this paper, we propose a novel method (named SPGA) to address the issues of limited positive samples and the imbalance of gradient contributions between easy and hard samples for robust visual tracking. In the proposed SPGA, we present an effective and efficient algorithm to generate positive samples in the feature space by using the Student's $t$-distribution. Moreover, to alleviate the imbalance problem between easy and hard samples, we introduce a gradient sensitive loss to harmonize the gradient contributions between easy and hard samples. Finally, we incorporate the proposed statistical positive sample generation algorithm and gradient sensitive loss into a baseline tracker, MDNet, to perform robust visual tracking (see Figure \ref{fig1} for qualitative tracking results).

The contributions of this paper can be summarized as follows:
\begin{itemize}
    \item We propose a novel statistical positive sample generation algorithm to effectively and efficiently enrich the diversity of positive training samples in the feature space to capture a variety of appearance changes over a temporal span.
    
    \item We propose to use a gradient sensitive loss to alleviate the imbalance problem of gradient contributions between easy and hard samples.
    
    \item Extensive experiments demonstrate the promising performance of the proposed method compared with state-of-the-art trackers on several visual tracking benchmarks.
\end{itemize}

\section{Related Work}
In this section, we mainly introduce some representative visual trackers and the related works on sample imbalance.

{\bfseries Visual Tracking. }
Correlation filters have gained considerable attention due to their computational efficiency and excellent performance. MOSSE \cite{MOSSE} firstly introduces the correlation filter, which encodes the target appearance through an adaptive filter by minimizing the output sum-of-squared error, to visual tracking. KCF \cite{KCF} formulates the kernelized correlation filters using circulant matrices.
DSST \cite{DSST} improves KCF by embedding scale estimation into correlation filters. SRDCF \cite{SRDCF}  employs a spatial regularization term to alleviate the boundary effects. Recently, with the development of deep learning, correlation filter based trackers take advantage of deep features instead of hand-crafted features \cite{DeepSRDCF,PCM1,DeepCFIAP} to significantly improve their performance. C-COT \cite{CCOT} proposes a continuous convolution operator to integrate multi-resolution deep features for efficient tracking.

In addition to integrating deep learning features into the framework of correlation filters, some works tend to design deep neural networks for tracking. SINT \cite{SINT} formulates visual tracking as a verification problem and trains a Siamese network to learn a metric for online target matching. SiamFC \cite{SiamFC} introduces cross-correlation into a fully convolutional Siamese network. RASNet \cite{rasnet} improves SiamFC by incorporating three attention modules to perform more discriminative metric learning. MDNet \cite{MDNet} trains a small network by using a multi-domain training strategy to obtain a generic representation of the target objects, achieving state-of-the-art performance. DLST \cite{DLST} decomposes the tracking problem into a localization task and a classification task, and it trains an individual network for each task. DAT \cite{DAT} proposes a reciprocative learning algorithm to exploit visual attention within the tracking framework. Meta-tracker \cite{Metatracker} incorporates meta-learning into visual trackers to quickly adapt the pre-trained network to the target in subsequent frames. These trackers advance the development of deep tracking models and achieve promising results on visual tracking benchmark datasets \cite{OTB2015, wu2013online, VOT2016}. 

{\bfseries Sample Imbalance. }
In visual tracking, the number of positive samples is limited while that of negative samples across the whole background is large. 
Meanwhile, a large portion of the training samples belong to easy samples, which dominate the computation of gradient during back-propagation in training a CNN based tracker. Recently, some solutions including data re-sampling \cite{ class-imbalance3} and cost sensitive loss \cite{vital, ghm, SiamFC} have been proposed to overcome the above mentioned problems. Among these solutions, SiamFC \cite{SiamFC} balances the loss of positive and negative samples in the score map for pre-training the Siamese network. MDNet \cite{MDNet} employs hard negative mining to alleviate the problem of data imbalance between easy and hard negative samples. VITAL \cite{vital} proposes a cost sensitive loss to reduce the influence of easy negative samples. Different from the above mentioned solutions, we propose a statistical positive sample generation algorithm to increase the number of positive samples. Furthermore, we present a gradient sensitive loss to harmonize the gradient contributions between easy and hard samples. 

\section{Proposed Method}
In this section, we firstly present the proposed statistical positive sample generation algorithm. Then, we introduce the proposed gradient sensitive loss which balances the gradient contributions between easy and hard samples. Finally, we describe the overall tracking pipeline of the proposed method.

\subsection{Statistical Positive Sample Generation}
\begin{figure*}[h]
	\setlength{\abovecaptionskip}{0pt}
	\setlength{\belowcaptionskip}{0pt}
    \centering
    \includegraphics[width = \linewidth, height = 0.29\linewidth]{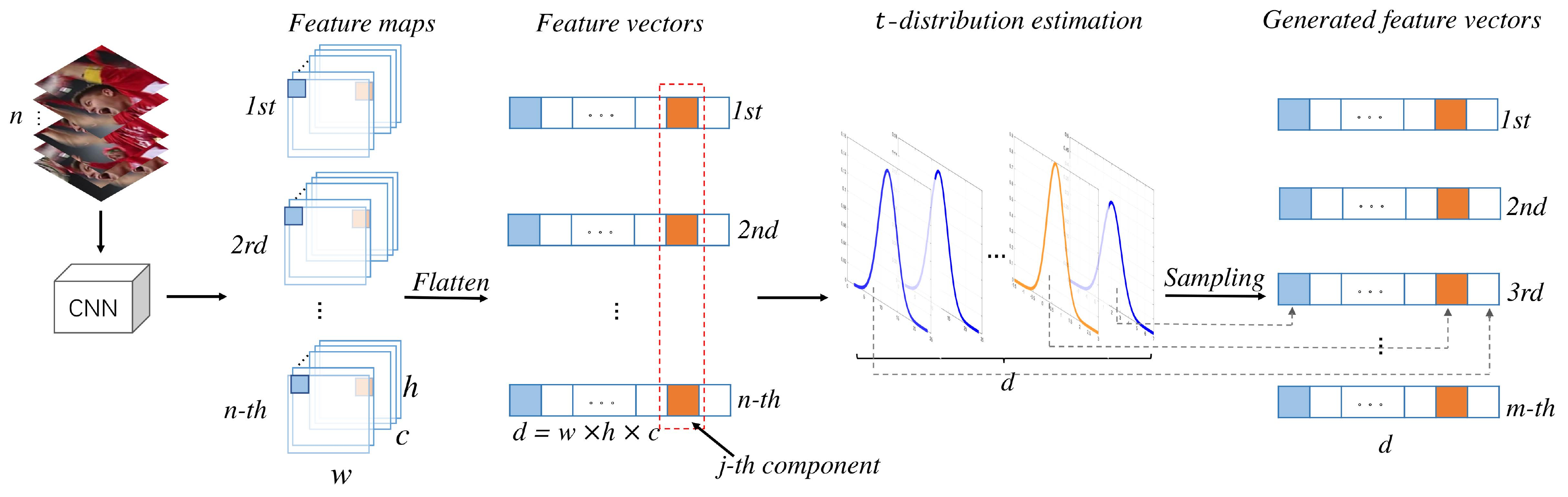}

    \caption{The pipeline of the proposed statistical positive sample generation algorithm. We first extract the feature maps of $n$ positive samples using CNN and then we flatten each of the feature maps into a vector. Then we adopt the Student $t$-distribution to estimate the distributions of all the components of each feature vector. Finally we generate $m$ feature vectors by randomly sampling based on the estimated feature distributions.}
    \label{fig:pipeline}

\end{figure*}
In visual tracking, positive samples are limited. Therefore, it is difficult for trackers to capture time-varying appearance changes of the target. Recently proposed trackers usually adopt GANs to augment the positive samples, which is time-consuming and unstable. In order to effectively and efficiently augment the positive samples, we propose a lightweight positive sample generation algorithm by using the Student's $t$-distribution \cite{t-distribution}. 
In probability and statistics, the Student's $t$-distribution can be used to estimate a confidence interval of the mean of a normally distributed population when the sample size is small and the population standard deviation is unknown. The pipeline of the proposed statistical positive sample generation algorithm is illustrated in Figure \ref{fig:pipeline}.

Specifically, given the original $n$ positive samples, we collect $n$ feature maps of these samples from the last convolutional layer of MDNet \cite{MDNet}. For each of the feature maps, we flatten it into a vector. We denote these feature vectors as ${\bm{\mathcal{F}} = \{\bm{F}_1, \bm{F}_2,..., \bm{F}_n\} \in \mathbb{R}^{n \times d}}$, where $n$ is the number of samples and $d$ indicates the length of each feature vector. 
We denote $f_i^j \in \bm{F}_i$ as the value for the $j$th component of the $i$th feature vector $\bm{F}_i$. 
We assume the $j$th components $f_{1:n}^j = \{f_1^j, f_2^j, ...,f_n^j\}$ of all the $n$ feature vectors are independent and they are distributed as a normal distribution $\mathcal{N}(\mu^j, \Sigma^j)$, i.e.,
\begin{equation}
    f_i^j \sim \mathcal{N}(\mu^j, \Sigma^j), i = 0,1,2,...,n,
\end{equation}\\
where $\mu^j$ and $\Sigma^j$ indicate the mean and variance of the normal distribution $\mathcal{N}(\mu^j, \Sigma^j)$. Thus, $f_{1:n}^j$ indicates $n$ observations from the normal distribution. 

Since the number $n$ of positive samples is small, we propose to use the Student's $t$-distribution to statistically estimate the confidence interval of $\mu^j$.
According to the Student's $t$-distribution, we can estimate the confidence interval $\bm{\mathcal{C}}^j$ of $\mu^j$ as follows:
\begin{equation}
    M^j - |\gamma _{\alpha /2}^j|\frac{S^j}{\sqrt{n}} < \mu ^j < M^j + |\gamma _{\alpha /2}^j| \frac{S^j}{\sqrt{n}},
\end{equation} \\
where $\gamma_{\alpha/2}^j$ indicates a pivotal quantity that follows the $t$-distribution with $n-1$ degrees of freedom and a two-sided significance level $\alpha$. $M^j$ and $S^j$ are the sample mean and sample variance of $f_{1:n}^j$, respectively. That is, 
\begin{equation}
    \mbox{Pr}(M^j - |\gamma _{\alpha /2}^j| \frac{S^j}{\sqrt{n}} < \mu ^j < M^j + |\gamma _{\alpha /2}^j |\frac{S^j}{\sqrt{n}})=1-\alpha,
\end{equation}\\
which means the probability that $\mu^j$ lies in $\bm{\mathcal{C}}^j$ is $1-\alpha$. Finally, the calculated confidence interval $\bm{\mathcal{C}}^j$ of $\mu^j$ is:

\begin{equation}
    \bm{\mathcal{C}}^j = [L^j,R^j] = [M^j - |\gamma _{\alpha /2}^j|\frac{S^j}{\sqrt{n}}, M^j + |\gamma _{\alpha /2}^j|\frac{S^j}{\sqrt{n}}],
\end{equation} \\
where $L^j$ and $ R^j$ respectively denote the lower bound and upper bound of $\bm{\mathcal{C}}^j$. In this paper, we use $\alpha = 0.05$ to effectively and efficiently estimate $\bm{\mathcal{C}}^j$. 

With the calculated intervals, we are able to generate a series of feature vectors $\bm{\mathcal{Z}} = \{\bm{z}_1, \bm{z}_2, ..., \bm{z}_m\} \in \mathbb{R}^{m \times d}$, where $m$ is the number of the generated feature vectors. The $j$th component of the $v$th generated feature vector $z_v^j \in \bm{\mathcal{Z}}$ is randomly sampled within the interval $[L^j, R^j]$. The final positive training samples are the combination of $\bm{\mathcal{F}}$ and $\bm{\mathcal{Z}}$. The proposed statistical positive sample generation algorithm guarantees that the generated feature vectors $\bm{\mathcal{Z}}$ are in the same domain as the original feature vectors $\bm{\mathcal{F}}$. Therefore, the proposed algorithm is lightweight and it is computationally efficient for positive sample generation.

\subsection{Gradient Sensitive Loss}
In visual tracking, a large portion of training samples belong to easy samples which contribute less discriminative information in training a binary classification network. In contrast, a minority of training samples are hard samples which provide much more discriminative information. Although a single easy sample contributes a little to the overall gradient, the summed gradient over a large number of easy samples can dominate the overall gradient, which makes the training process less effective. Re-weighting the samples is an effective way to down-weight the easy samples and emphasize the hard samples. 

In this paper, we propose a gradient sensitive loss, which is based on the cross entropy (CE) loss, to re-weight the training samples. In the CE loss, for a sample, let $p\in [0,1]$ denote the estimated probability prediected by a network
and $y \in \{0,1\}$ denote the ground-truth label of the sample. The CE loss can be formulated as:
\begin{equation}
    L(p,y) = -(y \cdot log(p) + (1-y) \cdot log(1-p)).
\end{equation}\\
Let $x$ be the direct output of the network and $p=sigmoid(x)$. Thus, we have the gradient with regard to $x$:
\begin{equation}
    \nabla_x L = 
    \begin{cases}
    p-1, & \mbox{if } y = 1 \\
    p,  & \mbox{if } y = 0\\
    \end{cases}.
\end{equation}\\
Then, the density of gradient of a sample can be formulated as:
\begin{equation}
    D(\nabla_x L) = \frac{1}{\epsilon}\sum_{k=1}^{N} \delta_\epsilon (\nabla_x L, \nabla_k L),
    \label{gradient}
\end{equation}\\
where $N$ is the number of training samples. $\nabla_k L$ is the gradient of the $k$th sample. And $\delta_\epsilon(\cdot , \cdot)$ is defined as:
\begin{equation}
    \delta_\epsilon(\nabla_x L, \nabla_k L) =
    \begin{cases}
    1,  &\mbox{if } \nabla_x L-\frac{\epsilon}{2} \leq \nabla_k L <  \nabla_x L+\frac{\epsilon}{2} \\
    0,  &\mbox{otherwise}\\
    \end{cases}.
\end{equation}\\
As a result, the gradient density $D(\nabla_x L)$ is calculated by the normalized number of samples residing in the region centered at $\nabla_x L$ with a length $\epsilon$. Due to the large number of easy samples, the gradient density of easy samples is usually large. In contrast, the gradient density of hard samples is usually small because of the limited number of hard samples. 

GHM \cite{ghm} is the first work in object detection which harmonizes the easy and hard samples based on the gradient density. It proposes to calculate the gradient density of a sample by using the distribution of the entire training set in object detection.
However, in visual tracking, the sample distribution between positive and negative samples is significantly different (i.e., the positive samples are highly overlapped with the target, while the negative samples are selected across the whole background). Therefore, we propose to calculate the gradient density of a sample solely based on the samples with the same class label (i.e., positive or negative).

Formally, according to the above observations, our proposed gradient sensitive loss is formulated by adding two weighting factors to the CE loss as:
\begin{equation}
    L_{GSL}(p,y) = -(y \cdot \lambda_{pos} \cdot log(p) + (1-y) \cdot \lambda_{neg} \cdot log(1-p)),
\end{equation}\\
where $\lambda_{pos}, \lambda_{neg}$ are the weighting factors which are calculated based on the gradient density to harmonize the gradient contributions of easy and hard samples:
\begin{equation}
    \lambda_{pos} = \frac{N_{pos}}{D_{pos}(\nabla_x L)},
\end{equation}\\
and 
\begin{equation}
    \lambda_{neg} = \frac{N_{neg}}{D_{neg}(\nabla_x L)},
\end{equation}\\
where $N_{pos}$ and $N_{neg}$ are the numbers of positive and negative samples. $D_{pos}(\nabla_x L)$ and $D_{neg}(\nabla_x L)$ denote the gradient densities which are respectively calculated based on the distribution of positive and negative samples according to Equation \ref{gradient}. Therefore, those samples with large gradient density (i.e., easy samples) are down-weighted and those samples with small gradient density (i.e., hard samples) are emphasized. By doing this, the gradient contributions of easy and hard samples are harmonized, which makes the classification network more discriminative.

\subsection{Online Tracking}
In this section, we describe how to incorporate the proposed statistical positive sample generation algorithm and gradient sensitive loss function into our baseline tracker (i.e., MDNet) to perform tracking. The details are as follows.

{\bfseries Model Initialization.} At the first frame, we randomly draw a number of samples. We replace the CE loss of MDNet with the proposed gradient sensitive loss and train the initial model using $I_1$ iterations.
As in MDNet, $n$ positive samples and $N_{neg}$ negative samples are collected in each iteration. We conduct the proposed statistical positive sample generation algorithm on the collected  positive samples to generate $m$ positive samples in the feature space. Therefore, the total number of positive samples $N_{pos}=n+m$. Finally, each mini-batch consists of $N_{pos}$ positive samples and $N_{neg}$ negative samples.

{\bfseries Online Detection.} When a new frame arrives, we randomly draw $N_1$ candidate samples around the predicted location in the previous frame. The candidate with the maximum target classification score is selected and then refined using bounding box regression.
 
{\bfseries Model Update.} As in MDNet, the model is updated every 10 frames or when potential tracking failures are detected. And we draw $N_2$ samples around the predicted target location. Similar to the model initialization, the model is updated using mini-batches with $N_{pos}$ positive and $N_{neg}$ negative samples for $I_2$ iterations.

\begin{figure}
\setlength{\abovecaptionskip}{5pt}
\setlength{\belowcaptionskip}{0pt}
\centering
    \includegraphics[width=0.49\linewidth, height = 0.35\linewidth]{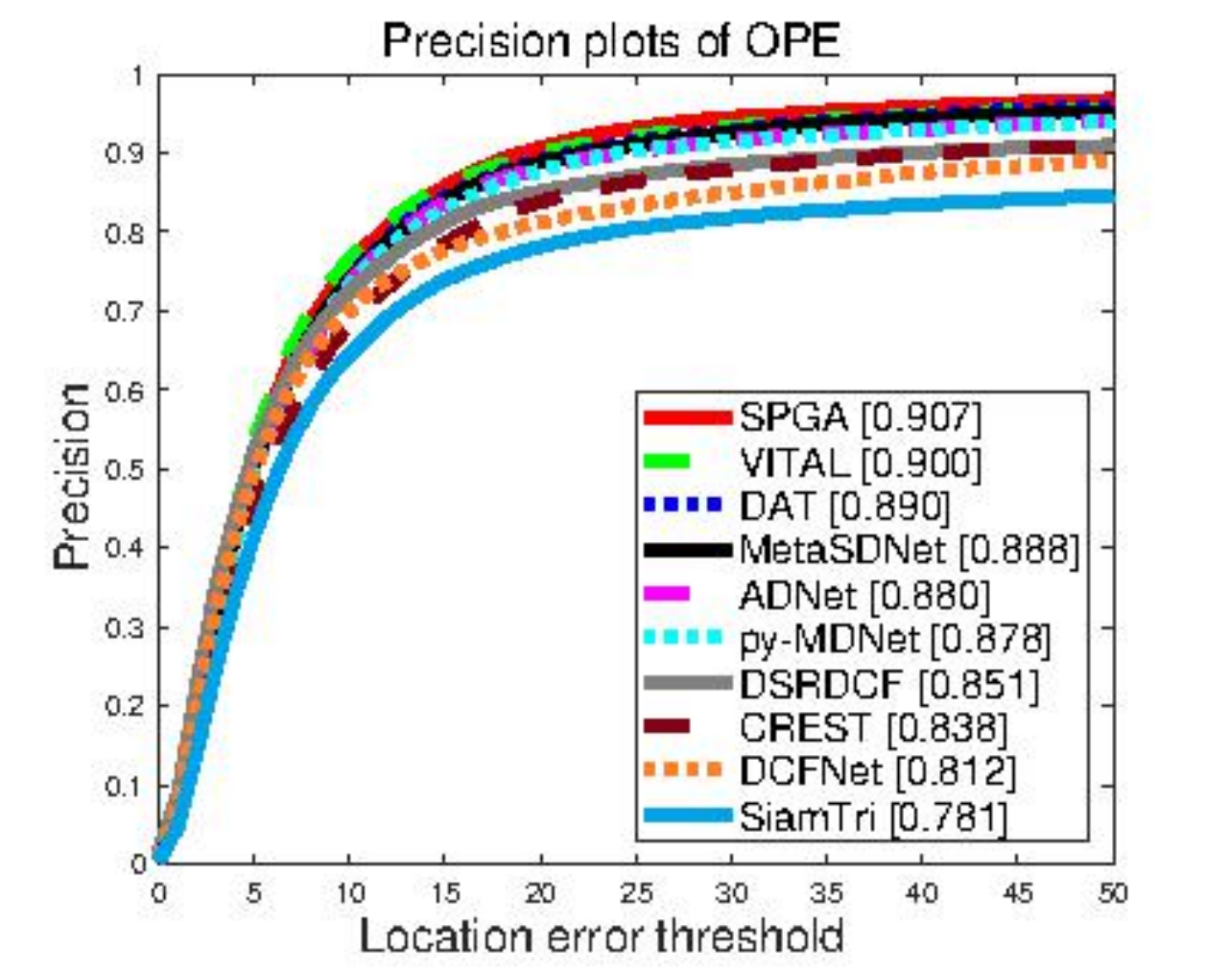}
    \includegraphics[width=0.49\linewidth, height = 0.35\linewidth]{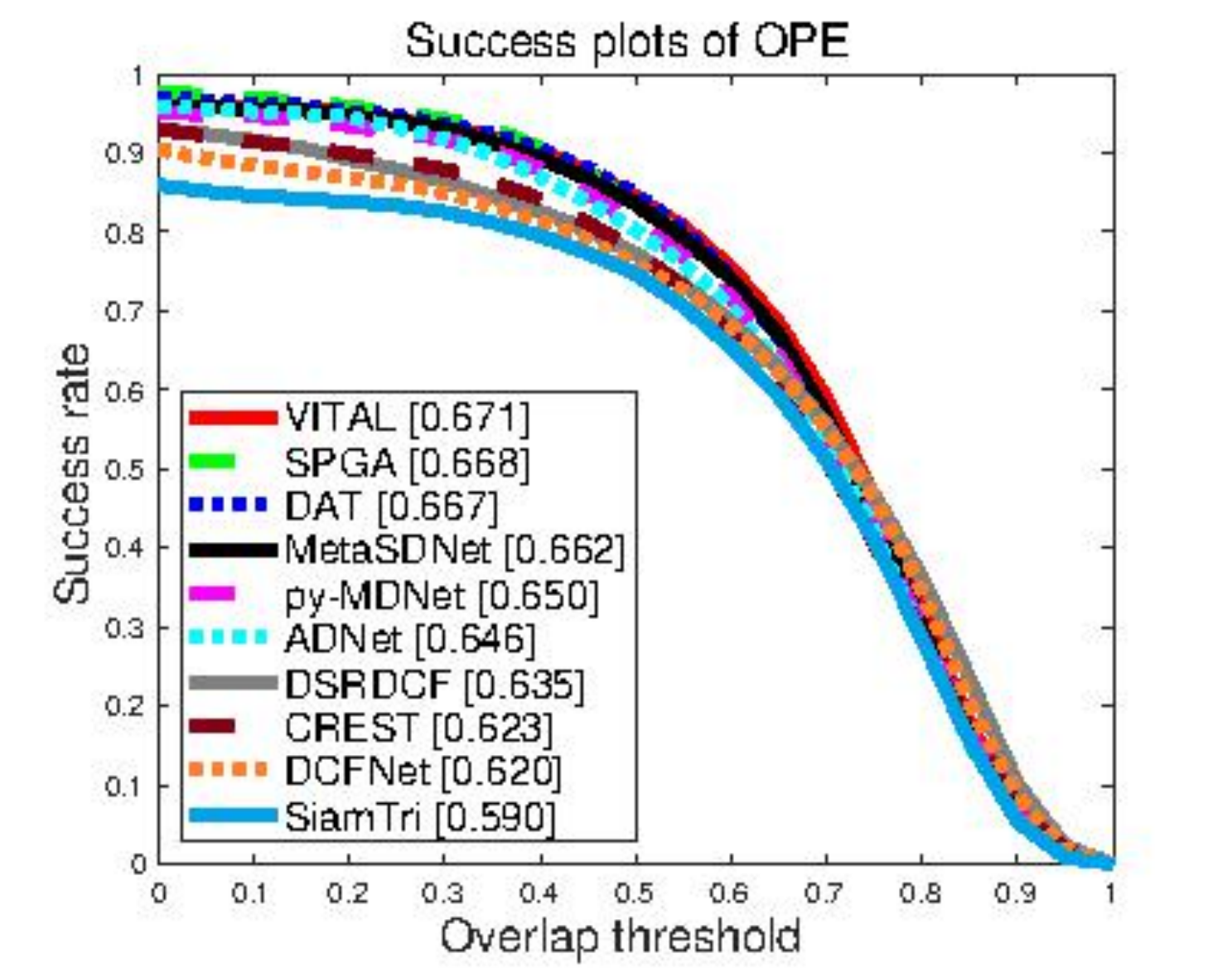}
  \caption{Precision and success plots on OTB100 obtained by the proposed SPGA and nine other state-of-the-art trackers.}
  \label{fig:OTB2015}
\end{figure}
\begin{figure*}
\setlength{\abovecaptionskip}{6pt}
\setlength{\belowcaptionskip}{0pt}
    \centering
    \includegraphics[width = 0.24\linewidth, height = 0.16\linewidth]{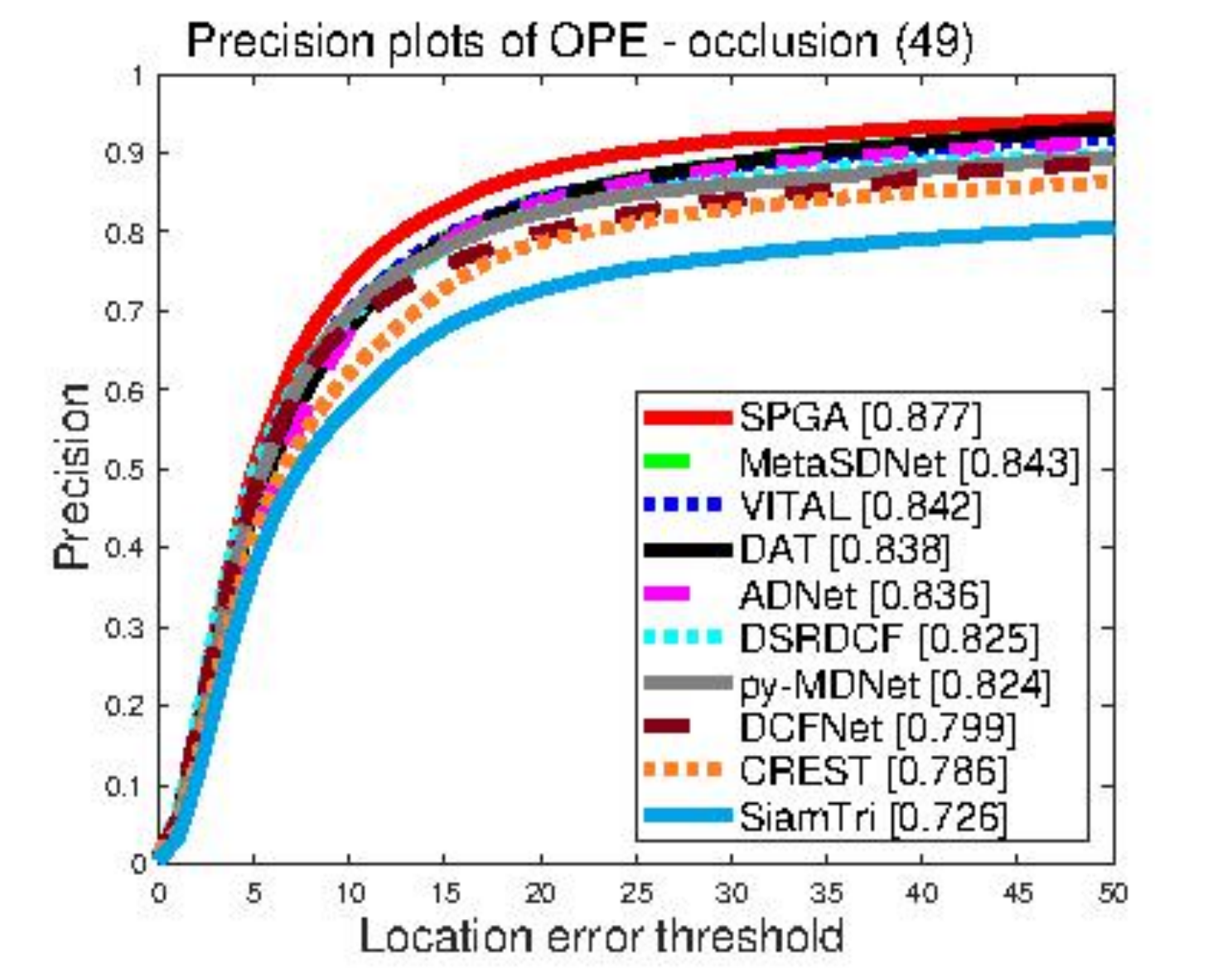}
    \includegraphics[width = 0.24\linewidth, height = 0.16\linewidth]{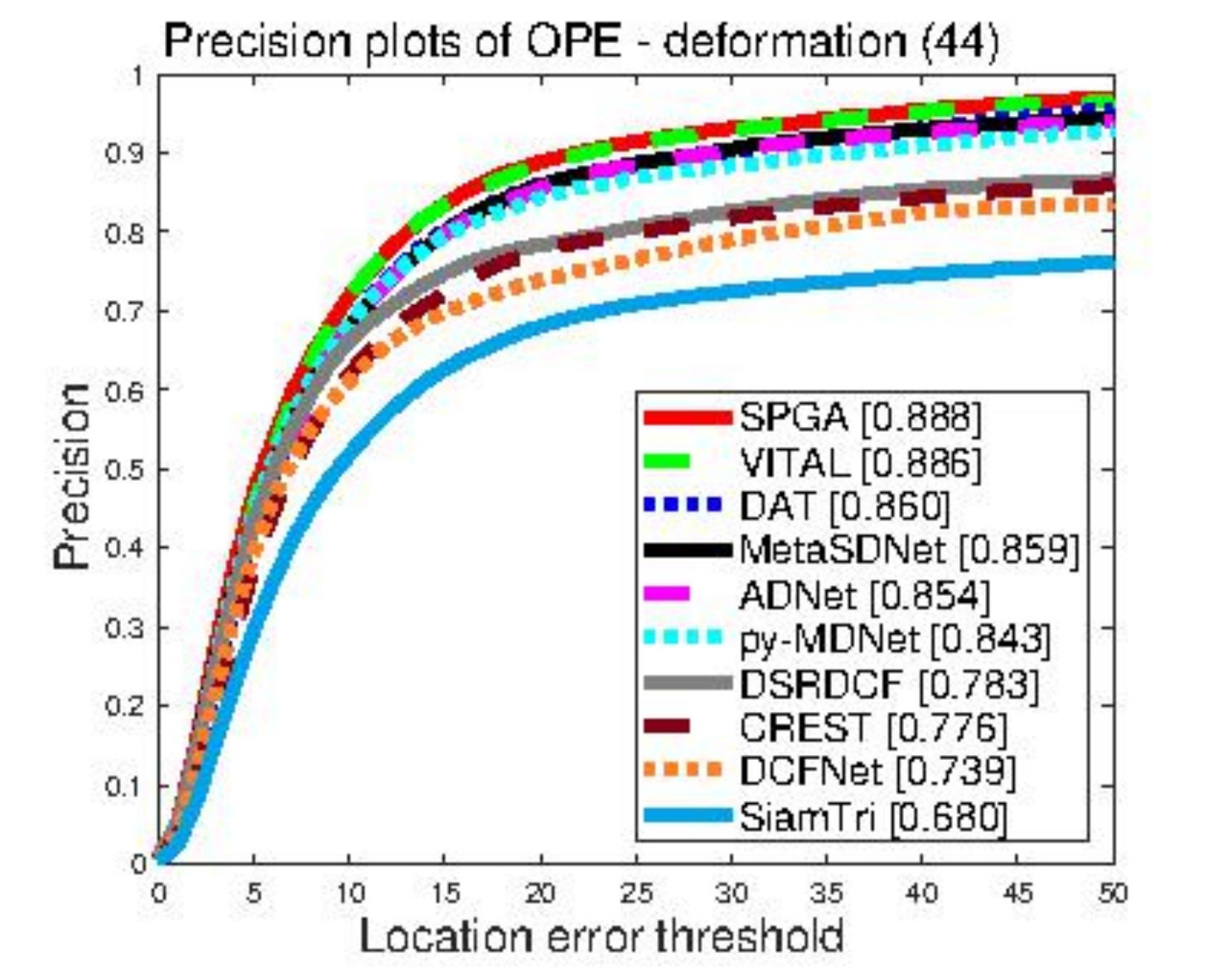}
    \includegraphics[width = 0.24\linewidth, height = 0.16\linewidth]{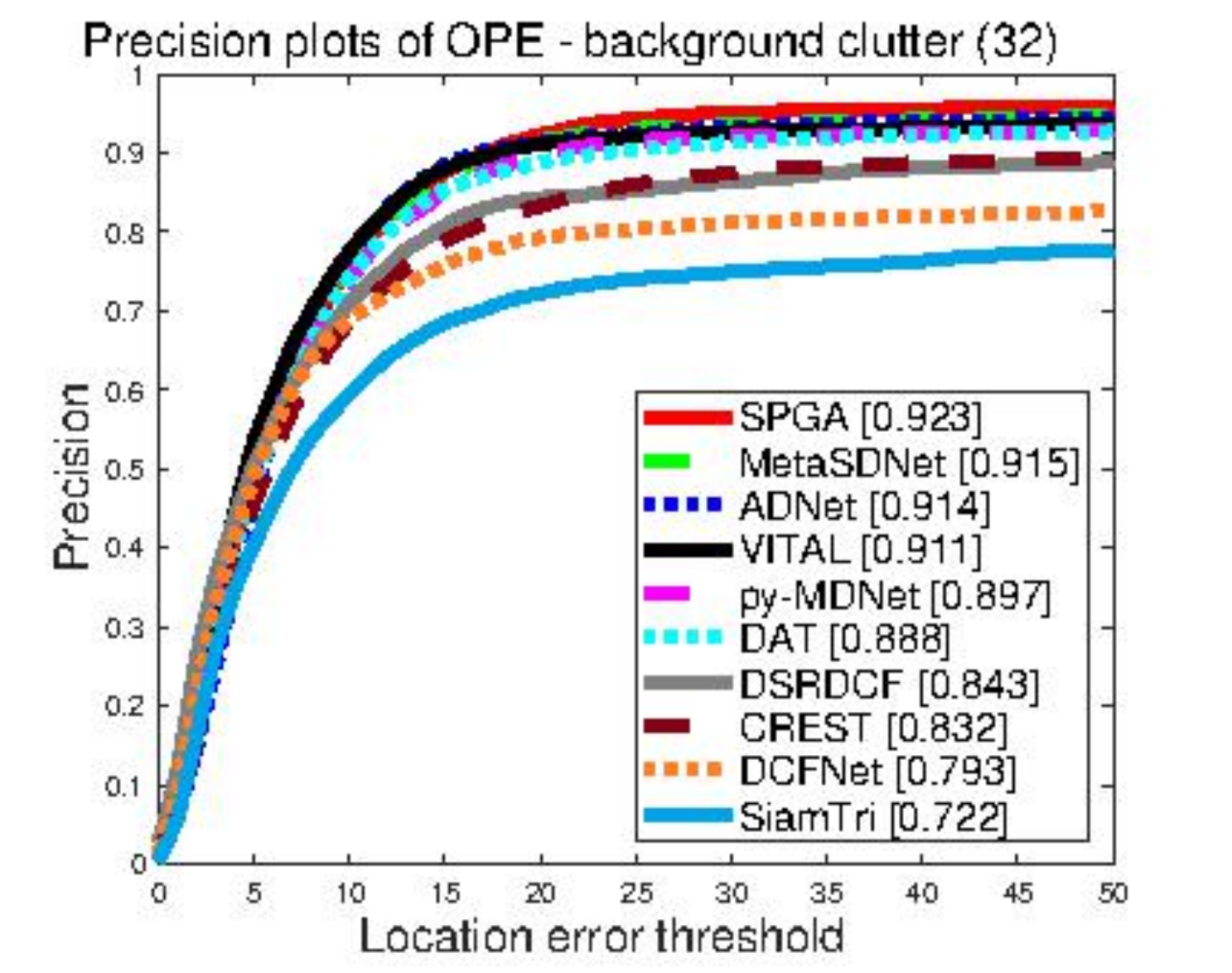}
    \includegraphics[width = 0.24\linewidth, height = 0.16\linewidth]{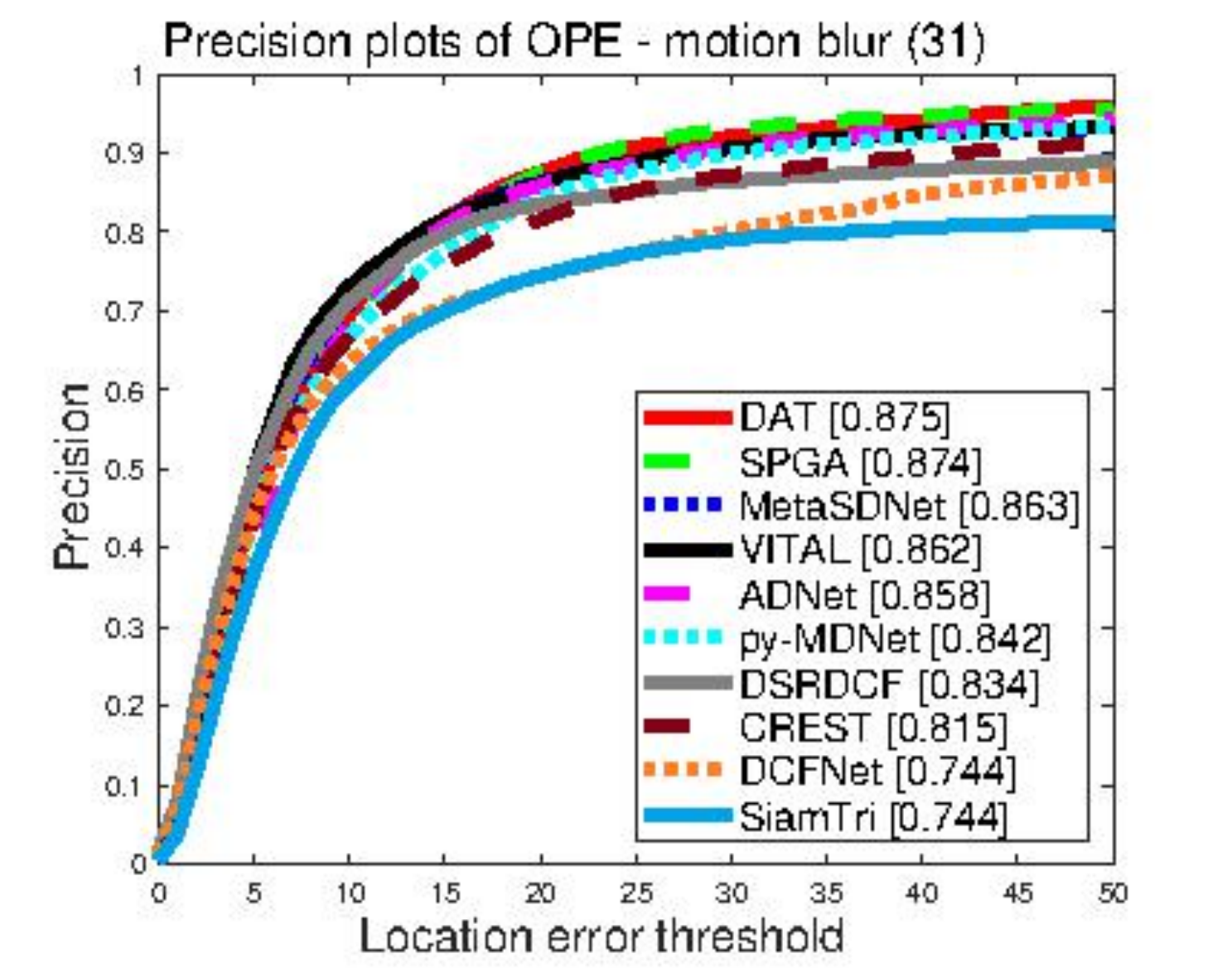}
    \includegraphics[width = 0.24\linewidth, height = 0.16\linewidth]{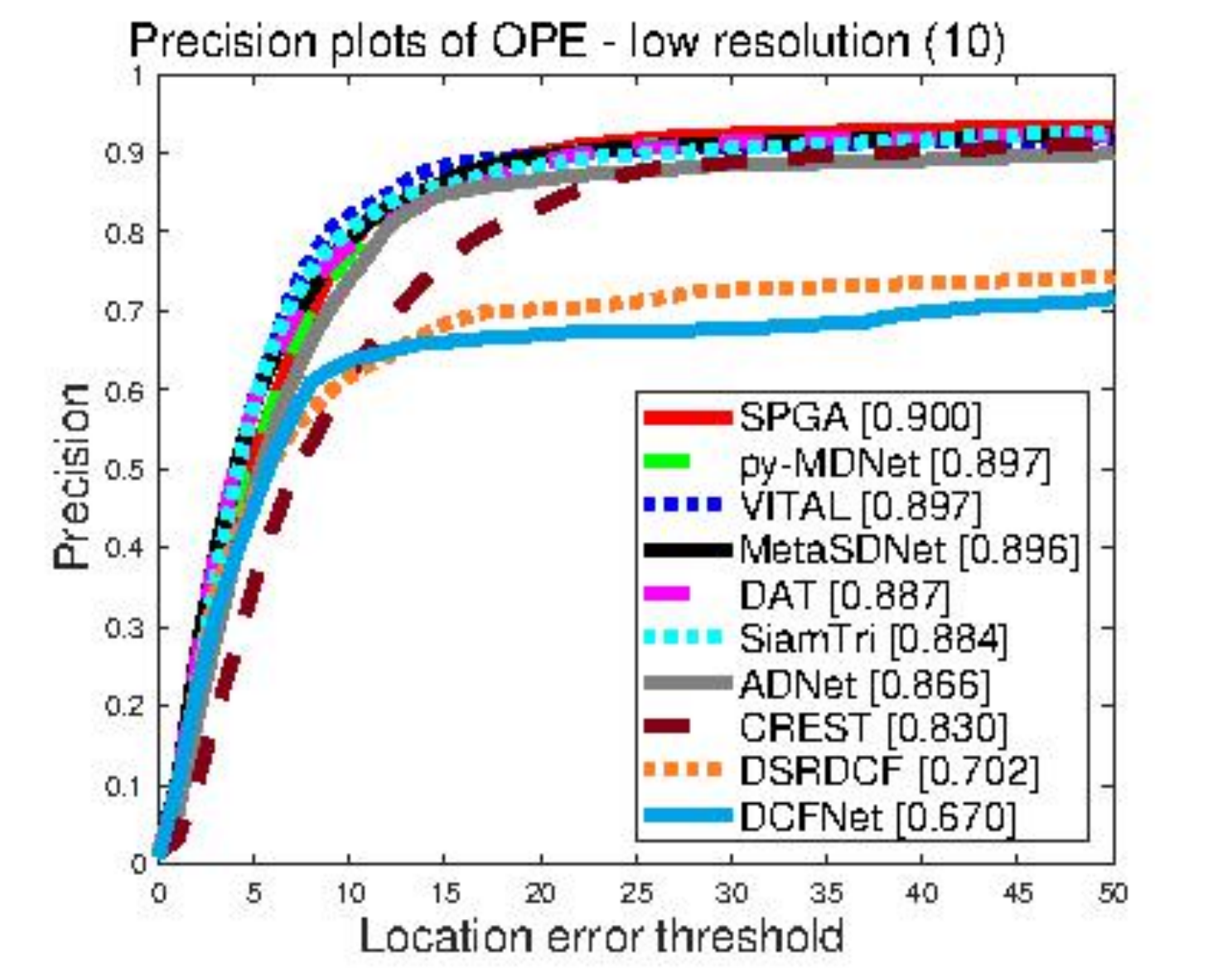}
    \includegraphics[width = 0.24\linewidth, height = 0.16\linewidth]{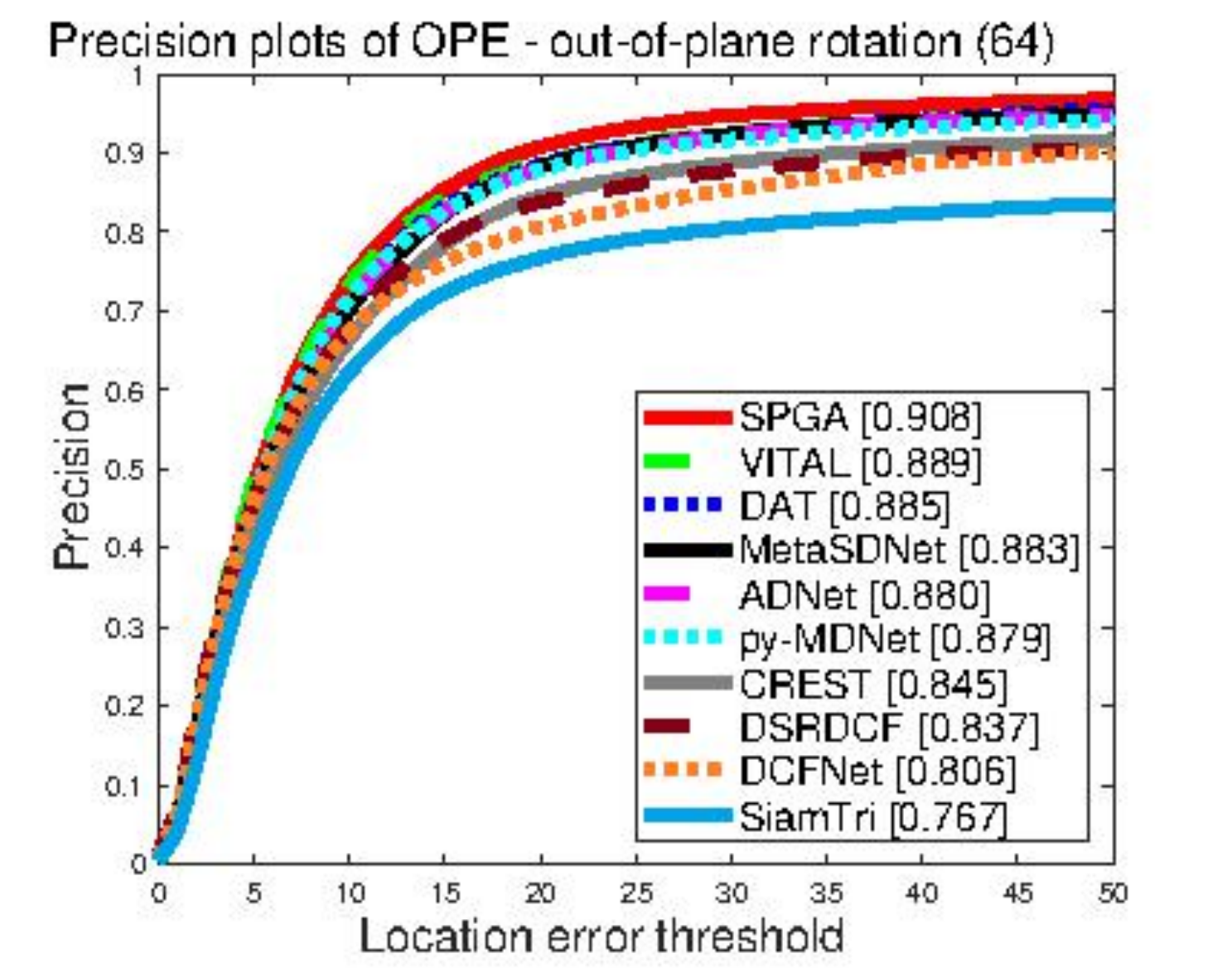}
    \includegraphics[width = 0.24\linewidth, height = 0.16\linewidth]{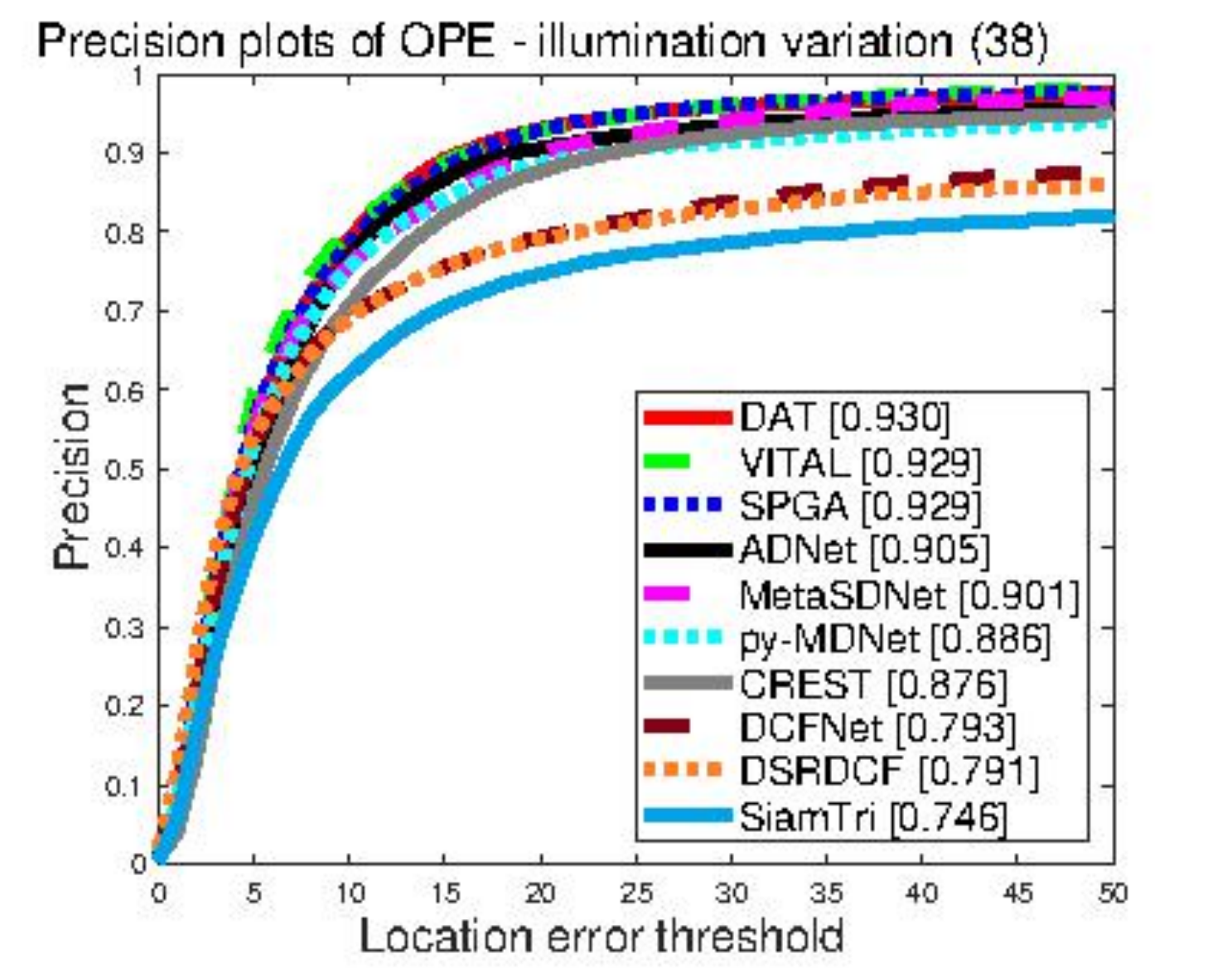}
    \includegraphics[width = 0.24\linewidth, height = 0.16\linewidth]{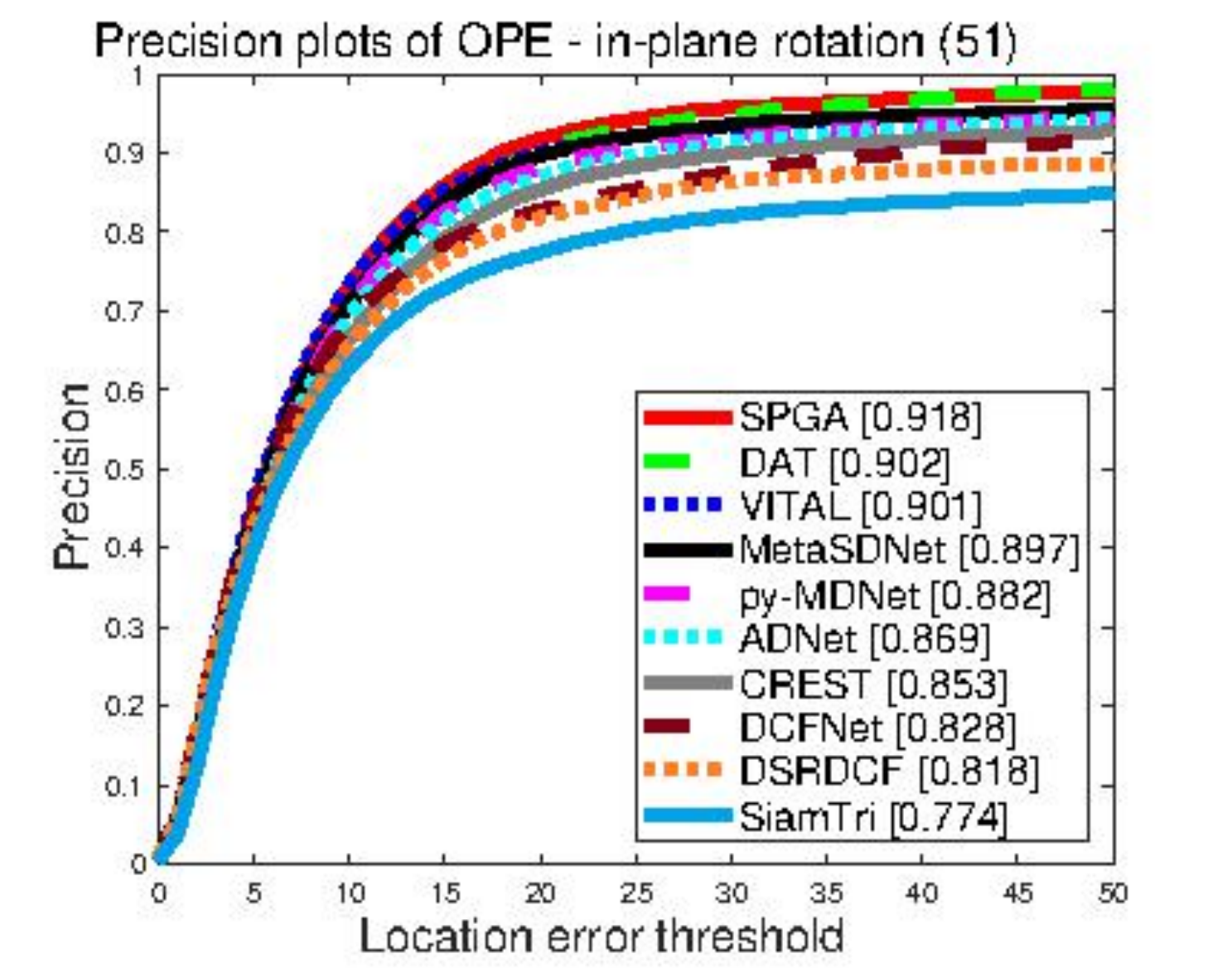}
    \caption{Precision plot on eight tracking challenge attributes, including occlusion, deformation, background clutter, motion blur, low resolution, out-of-plane rotation, illumination variation and in-plane-rotation on the OTB100 dataset.}
    \label{fig:tb15}
\end{figure*}

\section{Experiments}
We evaluate the proposed SPGA on three benchmark datasets including OTB50 \cite{wu2013online}, OTB100 \cite{OTB2015}, VOT2016 \cite{VOT2016}, and compare its performance with those obtained by several other state-of-the-art trackers. Our method is implemented in PyTorch using an Intel(R) Xeon(R) E5-2620 v4 CPU with 2.10GHz and an NVIDIA GeForce GTX1080 GPU. The proposed SPGA operates at around 2 FPS.

\subsection{Implementation Details}
We use the pre-trained MDNet \cite{MDNet} as our baseline tracker. MDNet uses three fixed convolutional layers and three learnable fully-connected layers. For the model initialization and the model update, we repectively train the fully-connected layers for $I_1 = 30$ and $I_2 = 10$ iterations with the same learning rates as in MDNet. In each iteration, we feed a mini-batch with $n = 32$ positive samples and $N_{neg} = 96$ negative samples into the network. Each feature map from the third convolutional layer is flattened into a vector with $d = 4608$ dimensions. We apply the proposed statistical positive sample generation algorithm to calculate the confidence interval of each component. Then $m = 64$ positive feature vectors are generated by uniformly sampling within the calculated intervals. Finally, the fully-connected layers are trained with $N_{pos} =96$ positive and $N_{neg} =96$ negative samples with the proposed gradient sensitive loss. For the online detection, the number of proposals $N_1$ is set to 256. For the model update, the number of samples $N_2$ is set to 250.

\subsection{Datasets and Evaluation Metrics}
The OTB50 and OTB100 datasets contain 50 and 100 sequences with various attributes, respectively.
We use the one-pass evaluation (OPE) with precision and success plots metrics on the OTB datasets. The precision plot shows the percentage of frames where the center of the tracking result are within 20 pixels away from the center of the ground-truth. The success plot is set to measure the overlap ratio between predicted and ground-truth bounding boxes. The VOT2016 dataset contains 60 sequences with various challenges. 
The performance is measured in terms of expected average overlap (EAO), accuracy (bounding box overlap ratio)  and robustness (number of failures), respectively.

\subsection{Ablation Study on OTB50 and OTB100}
We perform several ablation studies on OTB50 and OTB100 to investigate the effectiveness of the proposed SPGA. Table \ref{tab:ablation} gives the comparisons of the baseline tracker (i.e., MDNet) and its variants. 

{\bfseries Statistical Positive Sample Generation.} By introducing the proposed statistical positive sample generation algorithm into MDNet, denoted as MDNet\_SP, considerable precision gains ($+3.7\%$ on OTB50 and $+2.6\%$ on OTB100) are obtained. These gains are ascribed to the positive samples generated by the proposed statistical algorithm which enrich the diversity of the positive training samples. The average speed obtained by SPGA remains around 2 FPS. These results show the effectiveness and efficiency of the proposed statistical positive sample generation algorithm.

{\bfseries Gradient Sensitive Loss.} We investigate the impact of the proposed gradient sensitive loss by replacing the cross entropy loss of MDNet with the proposed gradient sensitive loss, denoted as MDNet\_GSL. As shown in Table \ref{tab:ablation}, MDNet\_GSL achieves better tracking performance than MDNet because GSL balances the gradient contributions between easy and hard samples. The balanced gradient contributions facilitate MDNet\_GSL to train a more discriminative classifier to effectively distinguish the target from background.

By integrating the above two components into MDNet, our SPGA achieves $+4.1\%$ and $+2.9\%$ precision gains respectively on OTB50 and OTB100, which outperforms MDNet by a large margin.
\begin{table}[h]
  \caption{Ablation study on OTB50 and OTB100. The results are presented in terms of OPE with precision. The best results are highlighted by bold.}
  \begin{tabular}{cccccl}
    \toprule
    dataset & MDNet & MDNet$\_$GSL & MDNet$\_$SP & SPGA \\
    \midrule
    OTB50 & 0.846 & 0.852 & 0.883 & $\bm{0.887}$ \\
    OTB100 & 0.878 & 0.880 & 0.904 & $\bm{0.907}$ \\
  \bottomrule
  \label{tab:ablation}
  \vspace{-2ex}
\end{tabular}
\end{table}

\subsection{Evaluation on OTB100}

In this section, the experiments on OTB100 are presented. We compare the proposed SPGA with several state-of-the-art trackers, including VITAL \cite{vital}, DAT \cite{DAT}, MetaSDNet \cite{Metatracker}, MDNet \cite{MDNet}, ADNet \cite{ADNet}, DeepSRDCF \cite{DeepSRDCF}, CREST \cite{crest}, DCFNet \cite{DCFNet}, SiamTri \cite{Triplet-tracker} and SiamFC \cite{SiamFC}. Among these trackers, VITAL, DAT and MetaSDNet are the state-of-the-art trackers based on MDNet. DCFNet, SiamTri and SiamFC are based on the Siamese network.

Figure \ref{fig:OTB2015} exhibits the precision and success plots on OTB100.
 For presentation clarity, we show the top 10 trackers. The proposed SPGA achieves the best performance in terms of precision and competitive overlap ratios compared with the top-ranked tracker (i.e., VITAL). Specifically, compared with the baseline MDNet tracker, SPGA achieves better results ($+2.9\%$ in precision and $+1.8\%$ in success rate), which can be attributed to the generated positive samples and the harmonized gradient contributions between easy and hard samples. 
SPGA achieves a slightly lower performance ($-0.03\%$) than VITAL in the success plots, but it yields better performance in the precision plots ($+0.07\%$). VITAL adopts GAN to augment positive samples to train the classifier, which makes it less efficient. Overall, the evaluation results demonstrate that our SPGA performs favorably against the state-of-the-art trackers on OTB100.

Moreover, we analyze the performance of the proposed SPGA under different attributes on OTB100. As shown in Figure \ref{fig:tb15}, SPGA outperforms the baseline MDNet tracker on the occlusion and in-plane rotation attributes by a significant margin ($+5.3\%$ and $+3.6\%$). This is because that the proposed statistical positive sample generation algorithm enriches the diversity of positive samples. Meanwhile, our SPGA achieves better performance in terms of background clutter and illumination variation ($+2.6\%$ and $+4.3\%$) compared with MDNet. This is because that the proposed gradient sensitive loss can effectively harmonize the gradient contributions between easy and hard samples, which facilitates a more discriminative classifier to distinguish the target from background. For other attributes, our SPGA performs the best in the precision plots, which demonstrates that SPGA is more effective in handling these challenges.

\begin{table}
  \caption{Comparisons with the state-of-the-art trackers on the VOT2016 dataset. The results are presented in terms of expected average overlap (EAO), accuracy (A) and failure (F). The best results are highlighted by bold.}
  \begin{tabular}{ccccccl}
    \toprule
    Trackers & SPGA &VITAL & MetaSDNet  &  SiamRN & MDNet \\
    \midrule
    EAO  & $\bm{0.32}$ & $\bm{0.32}$& 0.31 & 0.28 & 0.26 \\
    A & 0.51 & $\bm{0.56}$ &0.54 & 0.55 & 0.54  \\
    F  & $\bm{15.88}$ & 16.58 & 17.36 & 24.00 & 21.08 \\
  \bottomrule
  \label{tab:vot16}
  \vspace{-3ex}
\end{tabular}
\end{table}

\subsection{Evaluation on VOT2016}
For more thorough evaluations, we evaluate our SPGA on VOT2016 in comparison with four state-of-the-art trackers, including MetaSDNet \cite{Metatracker}, MDNet \cite{MDNet}, VITAL \cite{vital} and SiamRN \cite{VOT2016}.

Table \ref{tab:vot16} shows the evaluation results obtained by our SPGA and four other trackers. SPGA achieves competitive performance with VITAL under the EAO metric. However, VITAL uses GAN to augment positive samples, which only runs at around 0.5 FPS. While SPGA with the statistical positive sample generation algorithm runs at around 2 FPS. In addition, our SPGA outperforms the other competing trackers by generating positive samples and balancing the gradient contributions between easy and hard samples. Especially, our SPGA outperforms the baseline MDNet tracker by $23\%$ in terms of EAO,
demonstrating the effectiveness of the proposed statistical positive sample generation algorithm and gradient sensitive loss.

\section{Conclusion}
In this paper, we propose a robust tracking method by exploiting both the statistical positive sample generation algorithm and the gradient sensitive loss. Firstly, we propose a statistical positive sample generation algorithm to effectively generate a series of positive samples. It enriches the diversity of positive samples, leading to the better generalization ability of the proposed method. Then, we propose a gradient sensitive loss to alleviate the significant imbalance of gradient contributions between easy and hard samples, which contributes to a more discriminative classifer. Compared with the state-of-the-art trackers, the proposed SPGA achieves promising performance on both OTB and VOT benchmarks. 
\begin{acks}
This work is supported by the National Science Foundation of China (Grant Nos. U1605252, 61872307 and 61571379).
\end{acks}
\bibliographystyle{ACM-Reference-Format}
\bibliography{sample-base}

\appendix
\end{document}